\documentclass{article}

\usepackage{arxiv}

\usepackage{amsmath}
\usepackage[backend=biber,sorting=none]{biblatex}   
\usepackage{nicefrac}
\usepackage{listings} 
\usepackage[usenames,dvipsnames]{xcolor}
\usepackage{graphicx}
\usepackage{multicol,lipsum}
\usepackage{tabularx}

\usepackage[colorinlistoftodos,prependcaption,textsize=small]{todonotes}

\usepackage[utf8]{inputenc} 
\usepackage[T1]{fontenc}   
\usepackage{hyperref}      
\usepackage{url}           
\usepackage{booktabs}      
\usepackage{amsfonts}           
\usepackage{microtype}  

\usepackage{subcaption}
\DeclareCaptionFormat{custom}
{%
	\textbf{#1#2}\textit{\small #3}
}
\renewcommand{\vec}[1]{\boldsymbol{#1}}	

\usepackage{placeins}

\usepackage{draftwatermark}
\SetWatermarkText{PREPRINT}
\SetWatermarkScale{1}
\definecolor{wm}{gray}{0.95}
\SetWatermarkColor{wm}

\newcommand{\papertitle}{Eigen-informed NeuralODEs: Dealing with stability and convergence issues of NeuralODEs}
\newcommand{\paperkeywords}{NeuralODE, PhysicsAI, eigenvalue, eigenmode, stability, convergence, oscillation, stiffness, frequency, damping}

\newcommand{\myurl}[1]{{\small\url{#1}}}
\newcommand{\plotwidth}{{0.43 \textwidth}}

\newcommand{\reffig}[1]{Fig. \ref{fig:#1}}
\newcommand{\reftab}[1]{Tab. \ref{tab:#1}}
\newcommand{\refsec}[1]{Sec. \ref{sec:#1}}
\newcommand{\refequ}[1]{Equ. \ref{equ:#1}}

\newcommand{\seefig}[1]{(s. Fig. \ref{fig:#1})}
\newcommand{\seesec}[1]{(s. Sec. \ref{sec:#1})}

\usepackage{acronym}
\newacro{ANN}[ANN]{Artifical Neural Network}
\newacroplural{ANN}[ANNs]{Artifical Neural Networks}
\newacro{AD}[AD]{Automatic Differentiation}
\newacro{PINN}[PINN]{Physics-informed Neural Network}
\newacroplural{PINNs}[PINNs]{Physics-informed Neural Networks}
\newacro{ODE}[ODE]{ordinary differential equation}
\newacroplural{ODE}[ODEs]{ordinary differential equations}
\newacro{MSE}[MSE]{mean squared error}
\newacro{MAE}[MAE]{mean absolute error}
\newacro{VPO}[VPO]{Van der Pol oscillator}

\newacro{FRQ}[FRQ]{Frequency}
\newacro{STB}[STB]{Stability}
\newacro{OSC}[OSC]{Oscillation capability}
\newacro{DMP}[DMP]{Damping}
\newacro{STF}[STF]{Stiffness}
\newacro{SOL}[SOL]{Solution}

\renewcommand{\headeright}{Preprint}
\renewcommand{\undertitle}{Preprint}
\renewcommand{\shorttitle}{~}

\hypersetup{%
  pdftitle  = {\papertitle},
  pdfauthor = {Tobias Thummerer, Lars Mikelsons},
  pdfkeywords = {\paperkeywords}
}

\begin{document}

\title{\papertitle}
\author{%
	Tobias Thummerer \\
	Chair of Mechatronics\\
	Augsburg University \\
	\texttt{tobias.thummerer@uni-a.de}
	\And 
	Lars Mikelsons \\
	Chair of Mechatronics\\
	Augsburg University \\
	\texttt{lars.mikelsons@uni-a.de}  
}
\date{}
\maketitle

\begin{abstract}
Using vanilla NeuralODEs to model large and/or complex systems often fails due two reasons: Stability and convergence. 
NeuralODEs are capable of describing stable as well as instable dynamic systems. Selecting an appropriate numerical solver is not trivial, because NeuralODE properties change during training. If the NeuralODE becomes more stiff, a suboptimal solver may need to perform very small solver steps, which significantly slows down the training process. If the NeuralODE becomes to instable, the numerical solver might not be able to solve it at all, which causes the training process to terminate. Often, this is tackled by choosing a computational expensive solver that is robust to instable and stiff ODEs, but at the cost of a significantly decreased training performance. Our method on the other hand, allows to enforce ODE properties that fit a specific solver or application-related boundary conditions.
Concerning the convergence behavior, NeuralODEs often tend to run into local minima, especially if the system to be learned is highly dynamic and/or oscillating over multiple periods. Because of the vanishing gradient at a local minimum, the NeuralODE is often not capable of leaving it and converge to the right solution.
We present a technique to add knowledge of ODE properties based on eigenvalues - like (partly) stability, oscillation capability, frequency, damping and/or stiffness - to the training objective of a NeuralODE. We exemplify our method at a linear as well as a nonlinear system model and show, that the presented training process is far more robust against local minima, instabilities and sparse data samples and improves training convergence and performance.
\end{abstract}


\begin{multicols}{2}

\section{Introduction}
NeuralODEs describe the structural combination of an \ac{ANN} and an \ac{ODE} solver, where the \ac{ANN} functions as right-hand side of the \ac{ODE} \cite{Chen:2018}. This way, dynamic system models can be trained and simulated without learning the numerical solving process itself. This powerful combination lead to impressive results in different application fields \cite{Ramadhan:2020, Thummerer:2022}. If large and/or complex systems shall be modeled with NeuralODEs, the corresponding \ac{ANN} becomes deeper and wider, and two major challenges need to be faced: Stability and convergence. These two properties of (Neural)\acp{ODE} are often neglected in simple applications or handled in a very specific way, that might not be reusable in other applications.

\subsection{Stability}
There are countless definitions on the system property \emph{stability}, therefore this term is introduced in a few lines. In this paper, a system is considered \emph{stable}, if all eigenvalues of the system matrix $\vec{A}$ have a negative real value. If the system matrix is not constant, a system might be stable for specific locations in state and time, while it may be instable in others. Furthermore, the process of pushing instable eigenvalues from the right to the left half of the complex plane is referred to as \emph{stabilization}. A straight-forward training process for NeuralODEs looks as follows: The NeuralODE is solved like a common ODE, the gathered solution is compared against some reference data (loss function) and finally the \ac{ANN} parameters inside of the NeuralODE are adapted on basis of the loss function gradient, so that a better fit compared to the reference data is achieved. 

\end{multicols}
\twocolumn{}
A NeuralODE is considered \emph{solvable}, if the solver is able to scale all system eigenvalues into its stability region, without performing steps smaller than a (solver specific or user defined) minimal step size. Without further arrangements, for a randomly initialized NeuralODE - meaning the \ac{ANN} parameters are random values from an initialization routine like e.g. Glorot \cite{Glorot:2010} - there are no guarantees that the initialized NeuralODE is (efficiently) solvable by a given \ac{ODE} solver. The positions of the eigenvalues of the resulting system are simply not considered during the initial parameter selection. 

If an (often instable) NeuralODE can not be solved, there is a major issue resulting from that: If the solver terminates the solving process, no solution can be obtained, no loss can be computed and no parameter updates can be performed. This state can not be cured by a new solving attempt, because the parameter values and therefore the \ac{ODE} stability is unchanged. To prevent this, it is necessary to initialize NeuralODEs in a solvable configuration to obtain a solution that can be used for training. But even if the NeuralODE is initialized solvable and the target solution is known to be solvable, there is no guarantee that the parameter configurations during the optimization process don't lead to an unsolvable (too instable) system. So beside the solvable initialization, an active stabilization during training must be considered to obtain a robust training process. 

Stabilizing linear NeuralODEs by constraining weights in the cost function is discussed in \cite{Tuor:2020}. In \cite{Kang:2021}, also the stabilization of nonlinear systems is presented by constraining weights, so that a Lyapunov-stable solution is obtained. The method presented in this contribution considers a more generic approach focusing linear and nonlinear, stable and instable systems, as well as additional \ac{ODE} properties besides stability: Eigen-frequencies, damping, stiffness and oscillation capability. Further, eigenvalues of the system are not affected indirectly by constraining weights during optimization like in \cite{Tuor:2020} and \cite{Kang:2021}, but directly by locating them in a differentiable way and forcing them into specific locations. This way, also (partly) instable systems can be trained, like the \ac{VPO} in the later example.

\subsection{Convergence}
\begin{figure}[h!]
	\centering
	\includegraphics[width=\plotwidth]{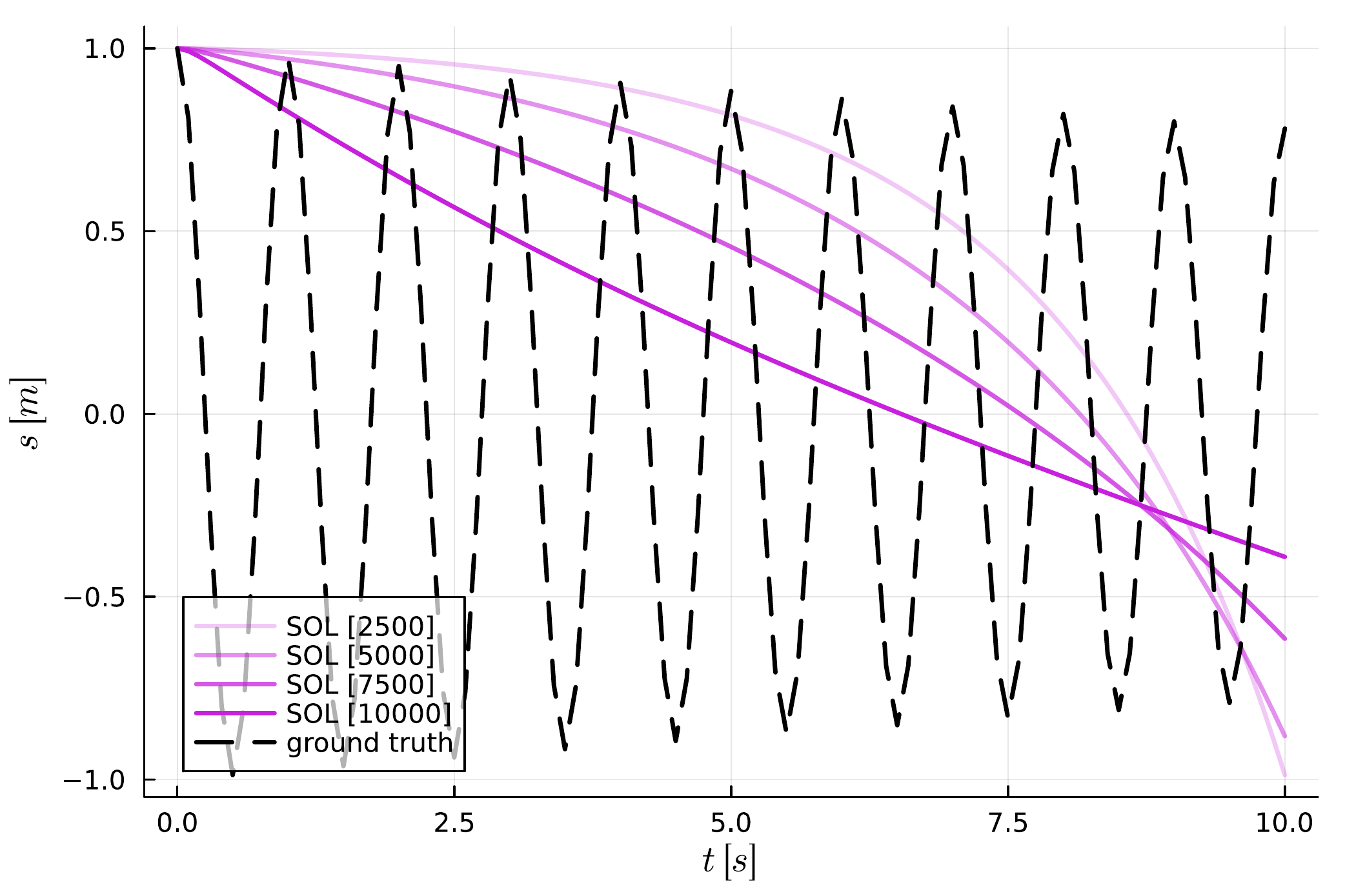}
	\caption{NeuralODE of an oscillating system, trained under conditions specified in \refsec{exp1}, after 2500, 5000, 7500 and 10000 steps. NeuralODE is suffering from local minimum. Because this local minimum is relative close to the optimum, leaving the local minimum is challenging because the loss function gradient vanishes.}
	\label{fig:oscmin}
\end{figure}

Even if the NeuralODE is (and stays) solvable, a common issue is the convergence to a local instead of the global minimum. This can be observed especially for oscillating systems, as is exemplified in \reffig{oscmin}. Instead of understanding a relative complicated oscillation, a much more simple average over the oscillation is learned. 
Without further arrangements, it is often impossible to train a NeuralODE to sufficiently describe an oscillating system over multiple periods or high dynamic system solely on base of a simple cost function defined on the \ac{ODE} solution. A common approach to meet this challenge is the so called \emph{growing horizon}, which starts with a small portion of the simulation horizon and successively increases this horizon with consecutive training convergence. Disadvantageous, this introduces multiple new hyperparameters\footnote{What is the horizon growing condition? What is the condition threshold? How much grows the horizon?}. Determination of these hyperparameters is not trivial. Especially for high frequency applications, the parameters are very sensitive. Also, the hyperparameters are very problem specific and chances of reusability in another application are low.

\section{Method}
The core idea of the presented method is to force the system eigenvalues into specific positions or ranges during training of the NeuralODE. This is achieved by a special loss function design on basis of the system eigenvalues. Inside the loss function, the method can be subdivided into three steps:
\begin{itemize}
	\item Gather the system matrix (jacobian) and compute the corresponding eigenvalues \seesec{eigvals},
	\item provide the eigenvalue sensitivities needed for the gradient over the loss function \seesec{eigvalssens} and
	\item rate the eigenvalue positions (compared to target positions) as part of a loss function \seesec{eigvalspos}. 
\end{itemize}

Because the evaluation of physical equations in the cost function of \acp{ANN} is known as \ac{PINN} \cite{Raissi:2019}, we want to pursue this naming convention by presenting \emph{eigen-informed} NeuralODEs, that evaluate eigenvalues (and/or -vectors) as part of their loss functions.

\subsection{Eigenvalues of the system}\label{sec:eigvals}
Eigenvalues are computed for the system matrix $\vec{A} = \frac{\partial \vec{\dot{x}}}{\partial \vec{x}}$, therefore the system state derivative vector $\vec{\dot{x}}$ is derived after the system state vector $\vec{x}$. For linear systems, this matrix is constant over the entire solving process, for nonlinear systems it is not (in general). As soon as a NeuralODE uses one or more nonlinear activation functions, it becomes a nonlinear system and the system matrix must be determined for every time instant. Determining this jacobian is not computational trivial, but often the system matrix was already calculated by another algorithm and can easily be reused. For example most of implicit solvers calculate and store the system jacobian for solving the next integrator step. In this case, the jacobian can be obtained without a computational effort worth mentioning. The other way round, also a computed jacobian for the application of this method can be shared with an implicit solver. If not available through another algorithm, the system matrix can be computed using \ac{AD} or sampled using finite differences. For some applications, even a symbolic jacobian may be available by default.

After obtaining the jacobian, the eigenvalues can be computed. There are multiple algorithms to approximate the eigenvalues for a given matrix $\vec{A}$. One of the most famous is the eigenvalue and -vector approximation via iterative QR-decomposition \cite{Francis:1962}. This algorithm may need many iterations to converge (for a tight convergence criterion), iteration count can be reduced by using different shift techniques or the algorithm extension \emph{deflation}. Using QR, the eigenvalues and -vectors can be computed.

\subsection{Sensitivities of eigenvalues}\label{sec:eigvalssens}
Computing sensitivities over the iterative QR algorithm using \ac{AD} is computational expensive, because additionally all algorithm operations must be performed on derivation level for every iteration. Further, the numerical precision on derivation level may decrease with larger iteration counts and an exact sensitivity computation is not guaranteed.

A better approach for sensitivity computation is to provide the sensitivities analytically over the entire iteration loop. The advantages are improved performance (the analytically expression needs only to be evaluated one time for an arbitrary number of iterations inside the algorithm) and improved numerical accuracy (no risk for numerical precision loss by iterating). 

Let $\vec{D} = diag(\vec{\lambda})$ be the diagonal matrix of eigenvalues $\vec{\lambda}$ and $\vec{U}$ a matrix that holds the corresponding eigenvectors in columns. For a function $\vec{D}, \vec{U} = eigen(\vec{A})$, that computes eigenvalues and -vectors, the sensitivities for forward mode differentiation $\vec{\dot{D}}$ and $\vec{\dot{U}}$ are provided in Equ. \ref{equ:forward1} and \ref{equ:forward2} based on \cite{Giles:2008}. 
\begin{equation}\label{equ:forward1}
\vec{\dot{D}} = \vec{I} \circ (\vec{U}^{-1} \cdot \vec{\dot{A}} \cdot \vec{U})
\end{equation}  
\begin{equation}\label{equ:forward2}
\vec{\dot{U}} = \vec{U} \cdot \left( \vec{F} \circ (\vec{U}^{-1} \cdot \vec{\dot{A}} \cdot \vec{U}) \right)
\end{equation} 
where $\cdot$ is the matrix product and $\circ$  is the Hadamard product (element-wise product). Further let $\vec{F}$ be defined as:
\begin{equation}
F_{i,j} = \left\{
\begin{matrix}
(\lambda_j - \lambda_i)^{-1} & \text{for $i \neq j$} \\
0 & \text{elsewhere} 
\end{matrix}
\right.
\end{equation} 
In analogy, the reverse mode sensitivities $\bar{\vec{A}}$ can be defined as in \cite{Giles:2008}:
\begin{equation}\label{equ:reverse}
\bar{\vec{A}} = \vec{U}^{-T} \cdot ( \bar{\vec{D}} + \vec{F} \circ (\vec{U}^T \cdot \bar{\vec{U}})) \cdot \vec{U}^T
\end{equation}
Both differentiation modes, forward and reverse, are implemented in our Julia package \emph{DifferentiableEigen.jl} (\url{https://github.com/ThummeTo/DifferentiableEigen.jl}).

\subsection{Eigenvalue positions}\label{sec:eigvalspos}
Inducing additional system knowledge in different forms into \acp{ANN} has been shown a good way to improve training convergence speed and reduce the amount of needed training data, compared to solving a task by a pure \ac{ANN} alone. The concept of NeuralODEs itself can be understood as the integration of an algorithm \emph{numerical integration} into an \ac{ANN}. On base of that, injecting one or more symbolic \acp{ODE} to obtain \emph{physics-enhanced} or \emph{hybrid} NeuralODEs further reduces the amount of training data and improves training convergence speed, because only the remaining, not modeled effects (or equations) need to be understood by the \ac{ANN} \cite{Thummerer:2021}. 

Continuing this strategy, also \emph{system properties} can be used as additional knowledge and can improve different aspects of NeuralODE training. This paper especially focuses on eigenmodes, meaning eigenvalues and eigenvectors, that can describe the following system attributes:
\begin{itemize}
	\item Stability (all eigenvalues on the left half of the complex plane)
	\item Oscillation capability (existence of conjugate complex eigenvalue pairs)
	\item Frequency (imaginary positions of conjugate complex eigenvalue pairs)
	\item Damping (positions of eigenvalues)
	\item Stiffness (ratio between largest and smallest, negative real positions of eigenvalues)
\end{itemize}
How to include knowledge of these system properties into a cost function of a NeuralODE, to obtain an \emph{eigen-informed} NeuralODE, is shown in the following subsections.

In the following, let $\vec{\lambda}(t)$ denote the eigenvalues of the corresponding system for the time instant $t$ and $\lambda_i(t)$ the $i$-th eigenvalue. The order of the eigenvalues is not important as long as it does not change\footnote{If the order of eigenvalues changes, e.g. because of function output with lexicographic sorting, eigenvalues must be tracked between time steps, so they can be uniquely identified.} over $t$. The real part of a complex eigenvalue $\lambda_i$ is denoted with $\Re(\lambda_i)$, the imaginary part with $\Im(\lambda_i)$.

For $a, b \in \mathbb{R}$ and $\vec{c} \in \mathbb{R}^n$, let $\epsilon(a, b)$ be an arbitrary function that values the deviation between $a$ and $b$. Further let $max(a, b)$ and $min(a, b)$ be the maximum/minimum function of two elements $a$ and $b$ and $max(\vec{c})$ and $min(\vec{c})$ the maximum/minimum element of $\vec{c}$.

\subsubsection{\ac{STB}}
Probably the most intuitively known system property, but often neglected during training of NeuralODEs, is \emph{stability}. Many physical systems are stable, and even instable mechanical or electrical systems relevant to practice often only contain a relative small instable subsystem. As already stated, NeuralODEs are \emph{not} stable by design. Stability can be achieved (and preserved) by forcing all real parts of the system eigenvalues to be negative. For instable systems, only the stable subset\footnote{This subset may vary over time, stable eigenvalues may become instable and the other way round.} of eigenvalues can be forced to the negative real half plane. A simple stability loss function $l_{STB}$ that forces this is described in the following:
\begin{equation}
	l_{STB}(t) = \sum_{i=1}^{|\vec{\lambda}|} \epsilon_{STB}(max(\Re(\lambda_i(t)),0), 0)
\end{equation}
The $max$ function with second argument $0$ ensures, that only eigenvalues with positive real value (instable) are taken into account. The error function $\epsilon_{STB}$ rates the deviation of the eigenvalue real part compared to $0$, where border-stable eigenvalues are located. Of course, other thresholds or ranges can be deployed if needed, for example to promote a stability margin instead of border-stability.

\subsubsection{\ac{OSC}}
Oscillation of a system can be determined by identifying conjugate complex eigenvalue pairs, meaning eigenvalue pairs with identical real parts, but negated imaginary parts. Eigenvalues can not appear with a non-zero imaginary part without a conjugate complex partner. As a consequence, to force two eigenvalues into an eigenvalue pair, it is necessary to start by synchronizing the real parts of the eigenvalues. For a given set $\vec{\Omega}$ of eigenvalue pairs $(\lambda_a, \lambda_b) \in \vec{\Omega}$ with $\lambda_a, \lambda_b \in \vec{\lambda}$, a cost function that forces oscillation can be defined as:
\begin{equation}
	l_{OSC}(t) = \sum_{(\lambda_a, \lambda_b) \in \vec{\Omega}} \epsilon_{OSC}(\Re(\lambda_a), \Re(\lambda_b))
\end{equation}
A point worth mentioning is, that it is not necessary to know exactly which eigenvalues should be paired. The knowledge \emph{that} eigenvalue pairs exist and the number of pairs is sufficient, because any two eigenvalues have the potential to establish a conjugate complex eigenvalue pair\footnote{This is because a NeuralODE can approximate any dynamic system, so any system eigenvector configuration, as long as the number of parameters is sufficient.}. If the order of eigenvalues changes (which is the case for many QR implementations), the eigenvalues need to be tracked, so that pairs can be maintained during training.

After laying the foundation for oscillation capability, it might be interesting to force a specific oscillation frequency (s. \refsec{frequency}) or damping (s. \refsec{damping}). 

\subsubsection{\ac{FRQ}}\label{sec:frequency}
Oscillation frequencies in dynamical systems are measured (and therefore known) in $Hz = \frac{1}{s}$. This information must be converted to eigenvalue positions in order to specify a loss function. As already stated, an oscillation is described by a pair of eigenvalues, sharing the same real part and a negated imaginary part. The frequency $f(\lambda_{a})$ of an eigenvalue $\lambda_{a}$ can be calculated by:
\begin{equation}
	f(\lambda_{a}) = \frac{ |\Im(\lambda_{a})| }{ 2 \cdot \pi }
\end{equation}
Based on that, a frequency based loss function may look as follows:
\begin{equation}\label{equ:lfreq}
	l_{FRQ}(t) = \sum_{(\lambda_a, \lambda_b) \in \vec{\Omega}} \epsilon_{FRQ}(f(\lambda_a), f(\lambda_b))
\end{equation}

\subsubsection{\ac{DMP}}\label{sec:damping}
Similar to the frequency, also the damping $\delta$ can be defined for an eigenvalue $\lambda_a$:
\begin{equation}
	\delta(\lambda_{a}) = \frac{-\Re(\lambda_{a})}{|\lambda_{a}|} = \frac{-\Re(\lambda_{a})}{\sqrt{\Re(\lambda_{a})^2 + \Im(\lambda_{a})^2 }}
\end{equation}
In analogy to \refequ{lfreq}, the damping loss function $l_{DMP}$ is defined straight forward:
\begin{equation}
	l_{DMP}(t) = \sum_{(\lambda_a, \lambda_b) \in \vec{\Omega}} \epsilon_{DMP}(\delta(\lambda_a), \delta(\lambda_b))
\end{equation}

\subsubsection{\ac{STF}}
Finally, also the stiffness ratio $\sigma$ of a system may be known or at least an upper or lower boundary. The stiffness loss function for stable systems can be defined as:
\begin{equation}
	l_{STF}(t) = \epsilon_{STF}\left(\frac{max(|\Re(\vec{\lambda})|)}{min(|\Re(\vec{\lambda})|)}, \sigma(t)\right)
\end{equation}
This stiffness definition basically only applies to stable systems, but for gradient determination it is important to provide a loss function that is defined for instable systems, too. Please note, that the error function $\epsilon_{STF}$ can be formulated not only to aim on a specific stiffness $\sigma$. Another approach is to allow a predefined stiffness corridor. Besides training the NeuralODE to a known stiffness or range, this feature can also be used to enhance simulation performance of the resulting NeuralODE. Common \ac{ODE} solver use the \ac{ODE} stiffness as criterion for the step size control, the stiffer the \ac{ODE} the more integration steps need to be performed\footnote{Basically, the step size scales the distance of the eigenvalues to the complex origin. To guarantee a stable solving process, all eigenvalues must be scaled into the solver's stability region.}. If the training goal is to solve the resulting NeuralODE with as least steps as possible, the training subject could be $\sigma=1$ to promote a fast simulating model.

\subsection{Training \& computational cost}\label{sec:cost}
Common optimizers for gradient based parameter optimization are designed to perform parameter changes on basis of a \emph{single} gradient. As soon as additional gradients are obtained, it is necessary to deploy a strategy on how to further progress with multiple gradients. Basically, there are three obvious ways to include one or more property loss functions along with the original loss function to the training process:
\begin{enumerate}
	\item Extending the original loss function, for example by adding other losses. This results in a single loss function gradient that can be passed to the optimizer.
	\item All gradients can be merged into one single gradient, which is used for an optimizer step. Merging techniques are for example: Sum or average of gradients or switching between gradients (based on a gradient criterion).
	\item If an optimizer robust to gradient changes is used (this is often the case for optimizers using momentum), all loss function gradients can be passed one after another to the optimizer to perform multiple optimization steps.
\end{enumerate}
A great feature of generating multiple gradients for NeuralODEs is, that multiple optimization directions are generated at a very low computational cost. This is not a matter of course and is only applicable, if essential parts of the gradient computation can be reused. This is the case, see the loss function $l$ gradient defined in \refequ{singlegrad} with \ac{ODE} solution $\vec{X}$ and \ac{ANN} parameters $\vec{\theta}$.
\begin{equation}\label{equ:singlegrad}
\frac{ \partial l(\vec{X}(\vec{\theta})) }{ \partial \vec{\theta} } = \frac{ \partial l(\vec{X}(\vec{\theta})) }{\partial \vec{X}(\vec{\theta})} \cdot \frac{\partial \vec{X}(\vec{\theta})}{\partial \vec{\theta}}
\end{equation}
The first part $\frac{ \partial l(\vec{X}(\vec{\theta})) }{\partial \vec{X}(\vec{\theta})}$ of the loss function gradient is computational cheap in general, it basically depends on the complexity of the used error function inside the loss function. The second factor, the jacobian $\frac{\partial \vec{X}(\vec{\theta})}{\partial \vec{\theta}}$ on the other hand is computational expensive, because it requires differentiation through the ODE solution $\vec{X}$, which requires differentiation through the \ac{ODE} solver. For every gradient that depends on the \ac{ODE} solution, $\frac{\partial \vec{X}(\vec{\theta})}{\partial \vec{\theta}}$ can be reused after being created a single time. As a result, the scalar loss function $l$ can be replaced by a vector loss function $\vec{l}$. Based on the loss function vector output, a loss function \emph{jacobian} can be obtained, with only little impact on the computational performance\footnote{In this paper's examples, computation time for the loss jacobian compared to the loss gradient increased by $\approx 1.5\%$ using forward mode \ac{AD}.}. This jacobian holds multiple optimization directions. The loss function jacobian is shown in \refequ{multigrad} and uses the same solution jacobian $\frac{\partial \vec{X}(\vec{\theta})}{\partial \vec{\theta}}$.
\begin{equation}\label{equ:multigrad}
\frac{ \partial \vec{l}(\vec{X}(\vec{\theta})) }{ \partial \vec{\theta} } = \frac{ \partial \vec{l}(\vec{X}(\vec{\theta})) }{\partial \vec{X}(\vec{\theta})} \cdot \frac{\partial \vec{X}(\vec{\theta})}{\partial \vec{\theta}}
\end{equation}
To conclude, for \emph{cheap} error functions like \ac{MSE} and similar, the computational cost for a single as well as for multiple gradients is driven by the cost of the jacobian over the solution $\frac{\partial \vec{X}(\vec{\theta})}{\partial \vec{\theta}}$. On the other hand, a cost function $\vec{l}_{\lambda}$ containing the eigenvalue operation is in general expensive\footnote{Besides the iterative nature of the QR algorithm, for sensitivity estimation through the eigenvalue determination also the inverse of $\vec{U}$ (s. Equ. \ref{equ:forward1}, \ref{equ:forward2} and \ref{equ:reverse}) is needed.}:
\begin{equation}
\frac{ \partial \vec{l}_{\lambda}(\vec{\lambda}(\vec{X}(\vec{\theta}))) }{ \partial \vec{\theta} } = \frac{ \partial \vec{l}_{\lambda}(\vec{\lambda}(\vec{X}(\vec{\theta}))) }{\partial \vec{\lambda}(\vec{X}(\vec{\theta}))} \cdot \frac{ \partial \vec{\lambda}(\vec{X}(\vec{\theta})) }{\partial \vec{X}(\vec{\theta})} \cdot \frac{\partial \vec{X}(\vec{\theta})}{\partial \vec{\theta}}
\end{equation}
Similar to the loss function defined on the solution, the jacobian $\frac{ \partial \vec{l}_{\lambda}(\vec{\lambda}) }{\partial \vec{\lambda}}$ basically depends on the used error function and is computationally cheap in general, whereas the eigenvalue jacobian $\frac{\partial \vec{\lambda}(\vec{X})}{\partial \vec{X}}$ needs derivation through the eigenvalue operations. Computational advantageous is, that as for the solution jacobian $\frac{\partial \vec{X}(\vec{\theta})}{\partial \vec{\theta}}$, the eigenvalue jacobian $\frac{\partial \vec{\lambda}(\vec{X})}{\partial \vec{X}}$ can be shared between all loss functions that consider eigenvalues and needs only be determined once.

To conclude, beside some computational savings, the number of time instances where the system matrix and eigenvalues are computed should be handled deliberately, because they may have significant impact on the overall training performance, dependent on the system dimensionality and complexity. As stated, if deploying implicit solvers, existing jacobians can be reused, resulting in major computational benefits.

\section{Experiments}
All experiments are using the \emph{Tsit5} solver \cite{Tsitouras:2011} for solving the Neural\ac{ODE} and the \emph{Adam} optimizer \cite{Diederik:2017} (default parameterization) for training the Neural\ac{ODE}. Some gradients are scaled to match the order of magnitude of the solution gradient: \ac{FRQ} (by $10^{1}$), \ac{STB} (by $10^{3}$), \ac{OSC} (by $10^{1}$) and \ac{STF} (by $10^{-1}$). The gradients are applied to the optimizer one after another, the Adam optimizer is robust to stochastic gradient changes.

Further, the examples are using a common, simple loss function to rate the \ac{ODE} \ac{SOL}:
\begin{equation}
	l_{SOL}(t) = \epsilon_{SOL}(x_1(t), \hat{x}_1(t))
\end{equation}
with $\hat{x}_1(t)$ being the ground truth data for the state $x_1(t)$ and $\epsilon_{SOL}(a,b)$ the \ac{MAE}. The corresponding program is written in the Julia programming language \cite{Bezanson:2017} using the Neural\ac{ODE} framework \emph{DiffEqFlux.jl} \cite{Rackauckas:2019}.

The potentials of the presented methodology is shown for three different use cases:
\begin{itemize}
	\item A weakly damped translational oscillator (linear system) to show the influence of different gradient setups (s. \refsec{exp1}),
	\item the same oscillator trained on sparse data, that does not fulfill the Nyquist-Shannon sampling theorem  (s. \refsec{exp2}) and
	\item the nonlinear \ac{VPO}, to show the applicability for nonlinear systems (s. \refsec{exp3}).
\end{itemize}
All \acp{ANN} inside the Neural\acp{ODE} are setup as described in \reftab{exp1}.
\begin{table}[h!] 
	\centering
	\caption{ANN layout for the examples.\label{tab:exp1}}
	\begin{tabularx}{\textwidth/2}{ccccr}
		\toprule
		Index & Type & Activation & Dimension & Parameters\\
		\midrule
		1 & Dense & tanh  & $2 \rightarrow 32$ & 96 \\
		2 & Dense & identity  & $32 \rightarrow 1$ & 33 \\
		\bottomrule
		~ & ~ & ~ & ~ & Sum: 129
	\end{tabularx}
\end{table}

\subsection{Linear system: Weakly-damped Oscillator}\label{sec:exp1}
As already stated in the introduction, especially oscillating systems are a challenging training task for NeuralODEs. By also applying a weak damping to the system, a longer simulation horizon must be considered, because training on a single oscillation period will not capture the damping effect. The state space equation of the translational spring-damper-oscillator with position $s$ and velocity $v$ is given in \refequ{linsys}. Please note, that this oscillator is a linear system\footnote{Training a linear system with a nonlinear NeuralODE is not the most reasonable approach. This was done to always keep the same NeuralODE topology for all examples.}. Training data is sampled equidistant with $10\,Hz$, simulation and training horizon are both $10\,s$. No batching is used, training is deployed on the entire \ac{ODE} solution.
\begin{equation}\label{equ:linsys}
	\vec{\dot{x}} = 
	\begin{bmatrix}
		\dot{x}_1\\
		\dot{x}_2
	\end{bmatrix} = 
	\begin{bmatrix}
	\dot{s}\\
	\dot{v}
	\end{bmatrix} = 
	\begin{bmatrix}
		v\\
		\frac{-c \cdot s - d \cdot v}{m}
	\end{bmatrix}
\end{equation}
with $c=25\,\frac{N}{m}$, $d=0.05\,\frac{N \cdot s}{m}$ and $m=1\,kg$.

\begin{figure}[h!]
	\centering
	\includegraphics[width=\plotwidth]{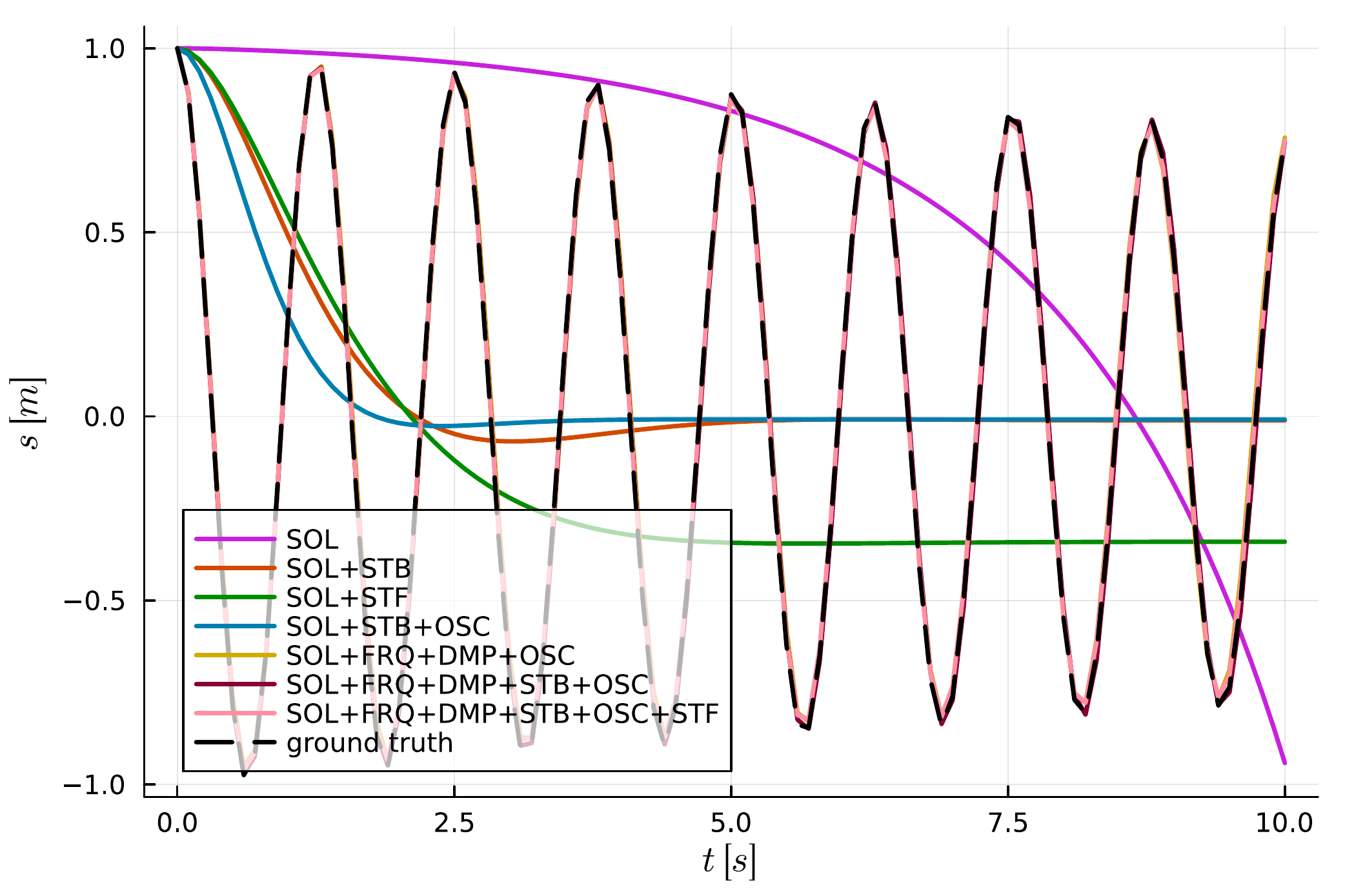}
	\caption{Comparison of the NeuralODE solutions (oscillator position), trained with different gradients. The gradient configurations containing FRQ and DMP (yellow, dark red, pink) lay almost exactly behind the ground truth (black-dashed).}
	\label{fig:transsol1}
\end{figure}

\begin{figure}[h!]
	\centering
	\includegraphics[width=\plotwidth]{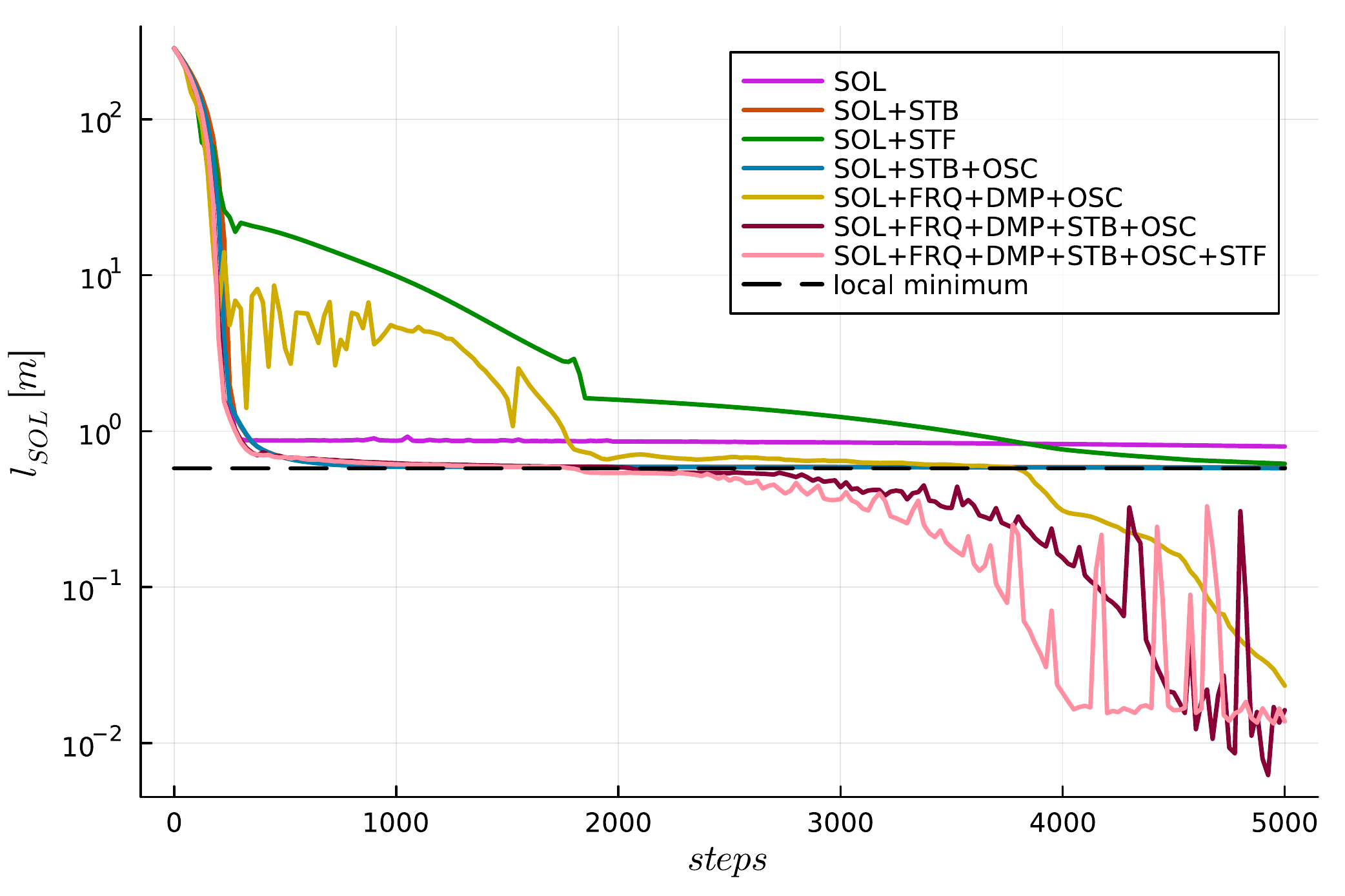}
	\caption{Comparison of the NeuralODE convergence behavior, trained with different gradients. Figure shows the loss defined on the ODE solution $l_{SOL}$. A strong local minimum is plotted black-dashed.}
	\label{fig:transconv1}
\end{figure}

\begin{figure}[h!]
	\centering
	\includegraphics[width=\plotwidth]{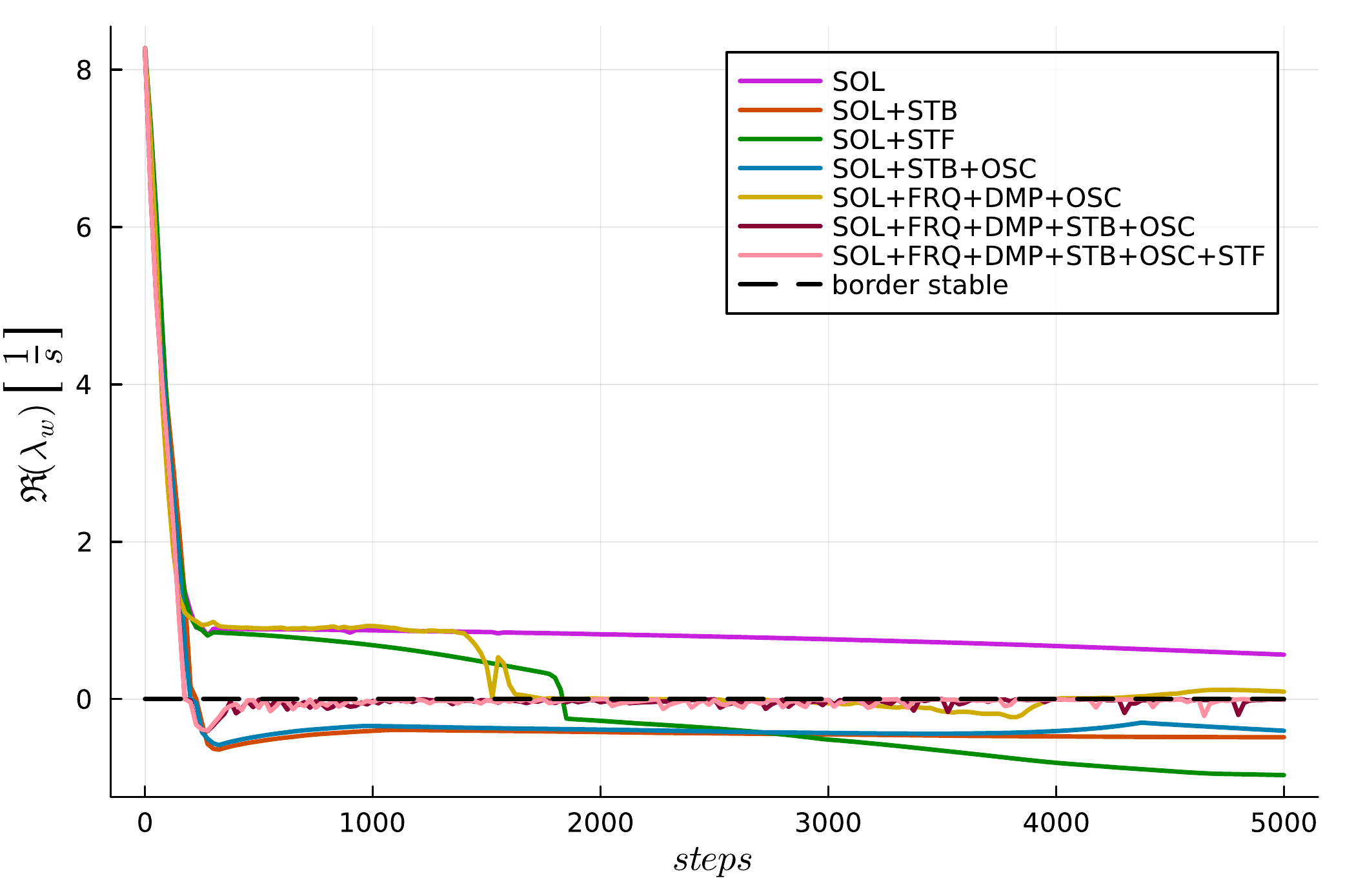}
	\caption{Comparison of the NeuralODE stability (rated by the maximum real value of the most instable eigenvalue $\lambda_w$), trained with different gradients. Stable systems are located beneath the border stable line (black-dashed).}
	\label{fig:transstab1}
\end{figure}

As can be seen in \reffig{transsol1}, the NeuralODE trained solely on its solution (SOL) is not capable to represent the linear system after a training of 5000 steps. By adding the stability gradient (STB), the solution converges in another local minimum, further adding the oscillation gradient (OCB) does not significantly improve the fit. Adding frequency (FRQ) and damping (DMP) information finally leads to a very good fit of the ground truth, even when removing the redundant\footnote{The stability information (negative real eigenvalues) is already contained in the damping gradient.} stability gradient. Adding the \ac{STF} gradient does not significantly influence training convergence compared to the corresponding gradient setups without \ac{STF} gradient. 
Similar, in \reffig{transconv1} the convergence behavior is presented. Basically, all gradient configurations run into the local minimum (black-dashed), except the configurations containing \ac{FRQ} and \ac{DMP} which are able to leave it before converging there.
Finally, the comparison of the solution stability (s. \reffig{transstab1}) shows the maximum real value over time of the most instable eigenvalue $\lambda_w$. As to expect, all gradient configurations containing the \ac{STB} gradient are stabilizing the system very fast during the first training steps and prevent the eigenvalues from leaving the stable half of the complex plain. 
\FloatBarrier

\subsection{Linear System: Insufficient data sampling frequency}\label{sec:exp2}
Another interesting use case is the training of a NeuralODE with data that does not fulfill the Nyquist–Shannon sampling theorem. To reproduce a signal that contains a maximum frequency of $b$ ($\frac{1}{s}$), data samples with a sampling distance $<\frac{1}{2b}$ ($s$) are needed. As a consequence, it is not possible to train a NeuralODE (or any other model) to reproduce high frequency effects that are not captured by a sufficient high data sampling rate. In practical applications, the data recording frequency is often limited by different factors, but eigenmodes of the system are often known. This information can be used in eigen-informed NeuralODEs.

\begin{figure}[h!]
	\centering
	\includegraphics[width=\plotwidth]{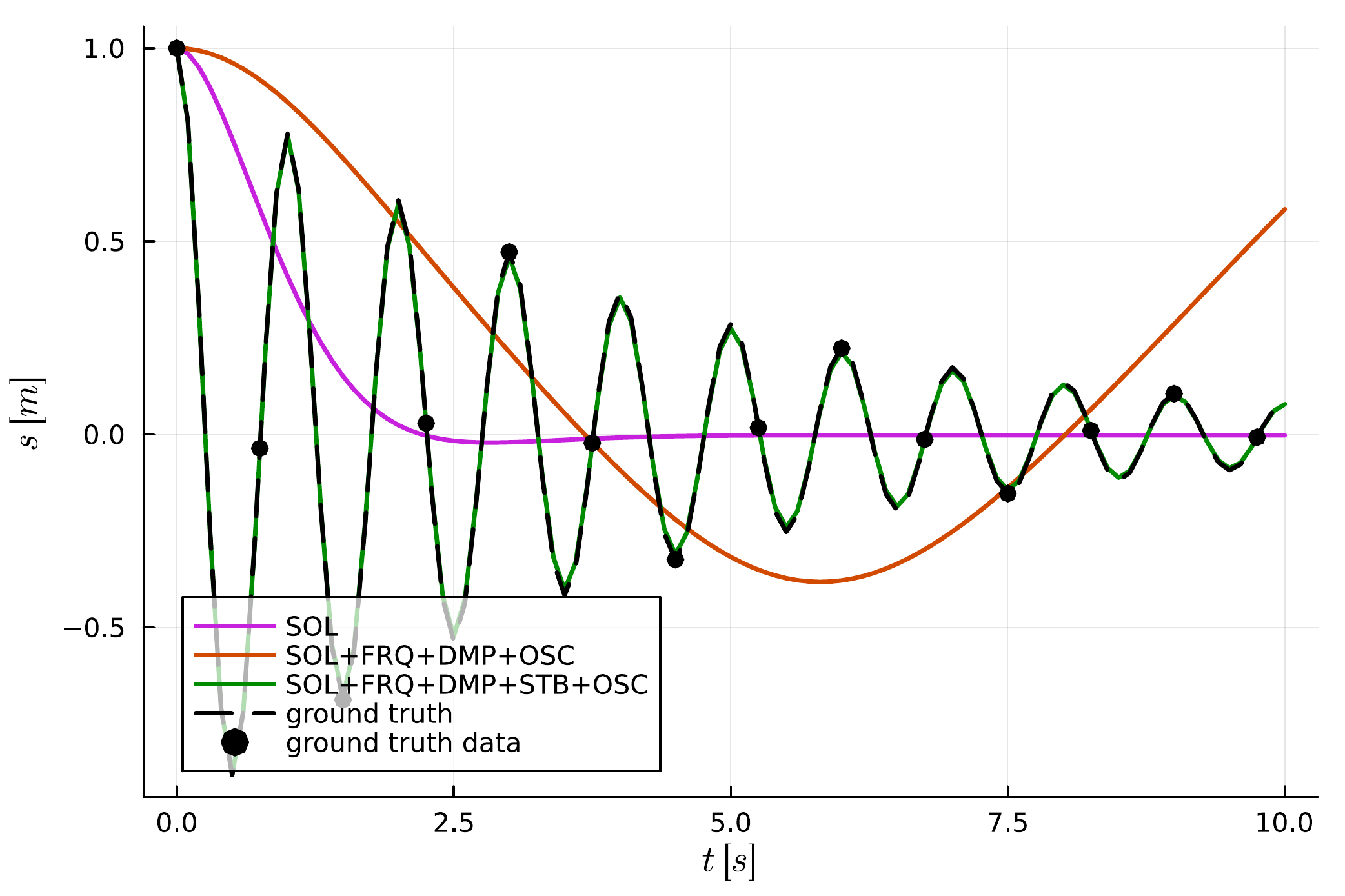}
	\caption{Comparison of the NeuralODE solutions (oscillator position), trained with different gradients. The gradient configuration SOL+FRQ+DMP+STB+OSC (green) lays almost exactly behind the ground truth (black-dashed).}
	\label{fig:transsol2}
\end{figure}

\begin{figure}[h!]
	\centering
	\includegraphics[width=\plotwidth]{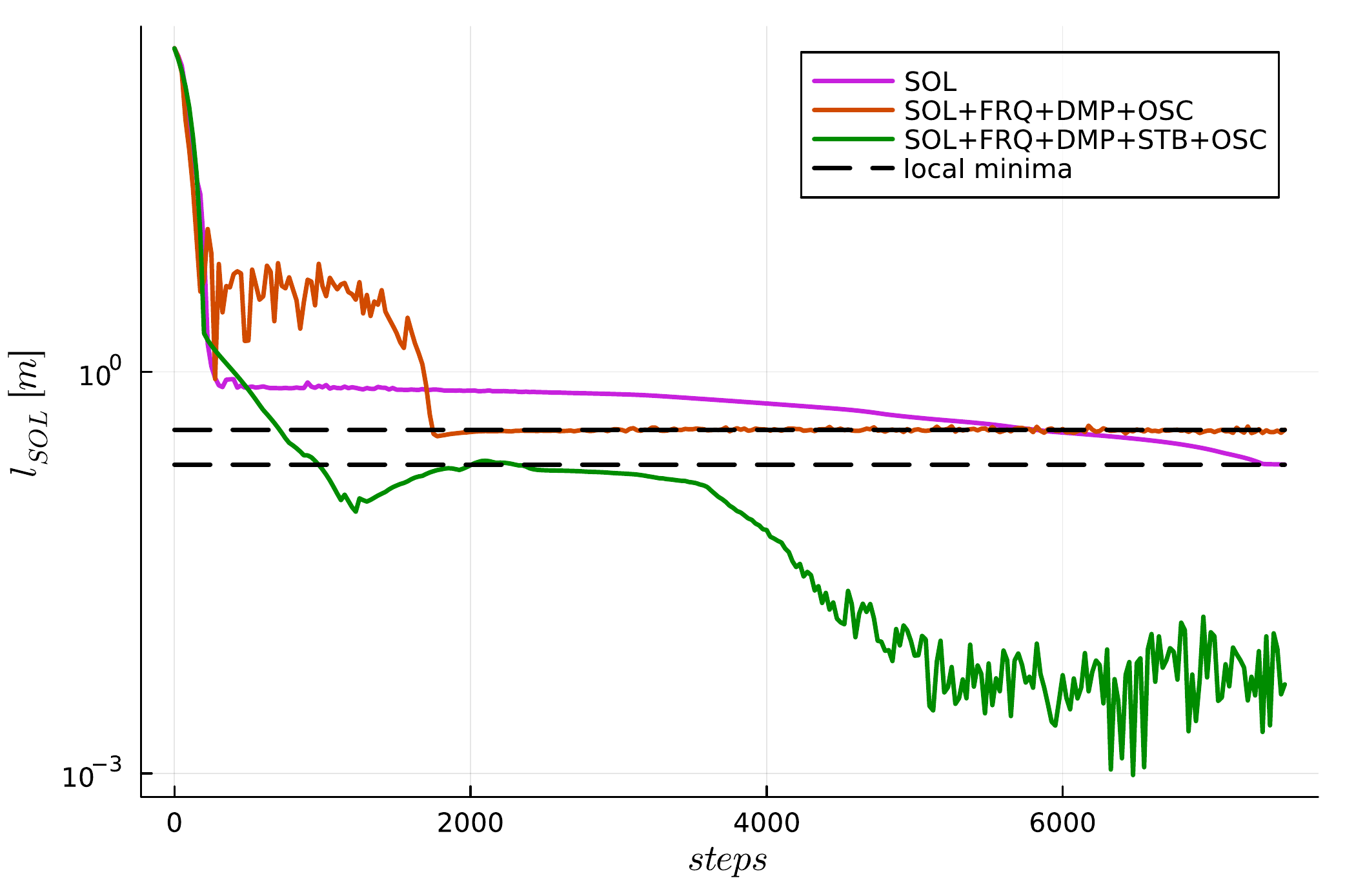}
	\caption{Comparison of the NeuralODE convergence behavior, trained with different gradients. Figure shows the loss defined on the ODE solution $l_{SOL}$. Two strong local minima are plotted black-dashed.}
	\label{fig:transconv2}
\end{figure}

\begin{figure}[h!]
	\centering
	\includegraphics[width=\plotwidth]{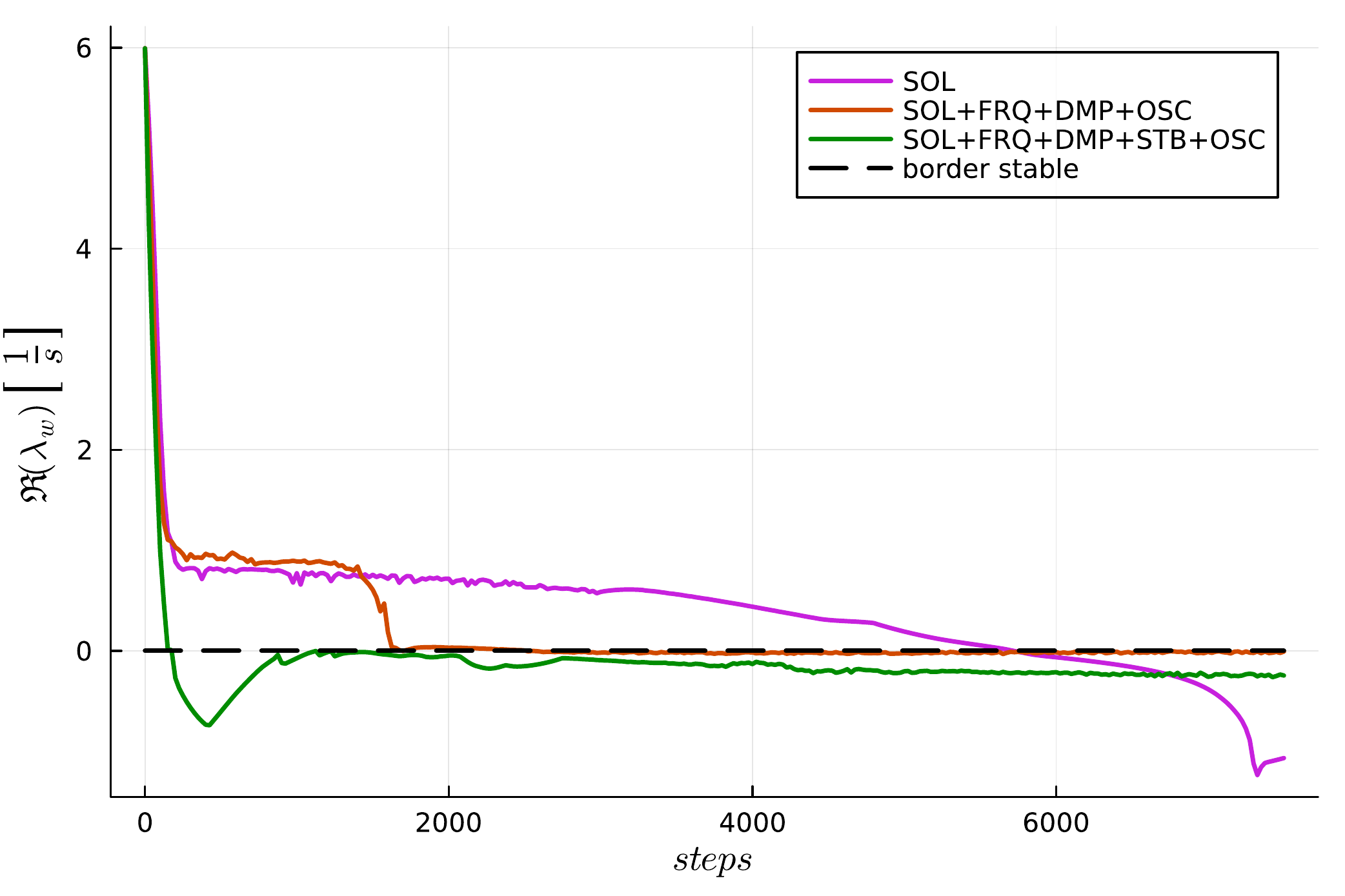}
	\caption{Comparison of the NeuralODE stability (rated by the maximum real value of the most instable eigenvalue $\lambda_w$), trained with different gradients. Stable systems are located beneath the border stable line (black-dashed).}
	\label{fig:transstab2}
\end{figure}

For this example, the same model as in \refsec{exp1} is used, but with a parameterization of $c=(2\pi \cdot f_{max})^2\,\frac{N}{m}$, $d=0.5\,\frac{N \cdot s}{m}$ and $m=1\,kg$. The maximum (and only) frequency of the oscillating system shall be $f_{max} = 1\,Hz$. The maximum distance between data samples $\Delta t_{max}$ can now be calculated using the Nyquist–Shannon sampling theorem:
\begin{equation}
	\Delta t_{max} = \frac{1}{2 \cdot f_{max}} = \frac{1}{2 \cdot 1\,\frac{1}{s}} = 0.5\,s
\end{equation}
Therefore, data sampling with $\Delta t > \Delta t_{max}$ will lead to an unrecoverable signal. For this example, we intentionally violate the theorem by picking $\Delta t = 0.75\,s > 0.5\,s = \Delta t_{max}$. As can be seen in \reffig{transsol2}, the resulting data sampling points are very sparse. As to expect, the NeuralODE based on the \ac{SOL} gradient is not able to replicate the signal and converges in a local minimum. Interestingly, also the NeuralODE additionally trained on the gradients \ac{FRQ}, \ac{DMP} and \ac{OSC} is not able to make a good prediction. Finally, only the gradient configuration adding the gradient \ac{STB} on top leads to a very good fit after 7,500 training steps.
As to expect, the convergence behavior (s. \reffig{transconv2}) of the gradient configuration SOL+FRQ+DMP+STB+OSC looks excellent compared to the other configuration. Both local minima (black-dashed) are passed, the lower minimum is shortly revisited at training step $\approx 2000$. The remaining configurations suffer from different local minima.
Regarding stability \seefig{transstab2}, especially the configuration including the STB gradient is stabilized very fast and kept stable for the entire training process. Also the configuration SOL+FRQ+DMP+OSC (orange) is trained border-stable. After training step $\approx 6000$, also the gradient configuration SOL becomes stable by converging in a local minimum.
\FloatBarrier

\subsection{Nonlinear system: \acl{VPO} (\acs{VPO})}\label{sec:exp3}
Finally, also a partly instable and nonlinear system is observed: The \ac{VPO}. The nonlinear system is given by the well-known state space equation in \refequ{nonlinsys}. Training data is sampled equidistant with $3\,Hz$, simulation and training horizon are both $30\,s$. No batching is used, training is deployed on the entire \ac{ODE} solution.
\begin{equation}\label{equ:nonlinsys}
\vec{\dot{x}} = 
\begin{bmatrix}
\dot{\nu}\\
\ddot{\nu}
\end{bmatrix} = 
\begin{bmatrix}
\dot{\nu}\\
\mu \cdot (1- \nu^2) \cdot \dot{\nu} - \nu
\end{bmatrix}
\end{equation} 
with $\mu = 2$.

\begin{figure}[h!]
	\centering
	\includegraphics[width=\plotwidth]{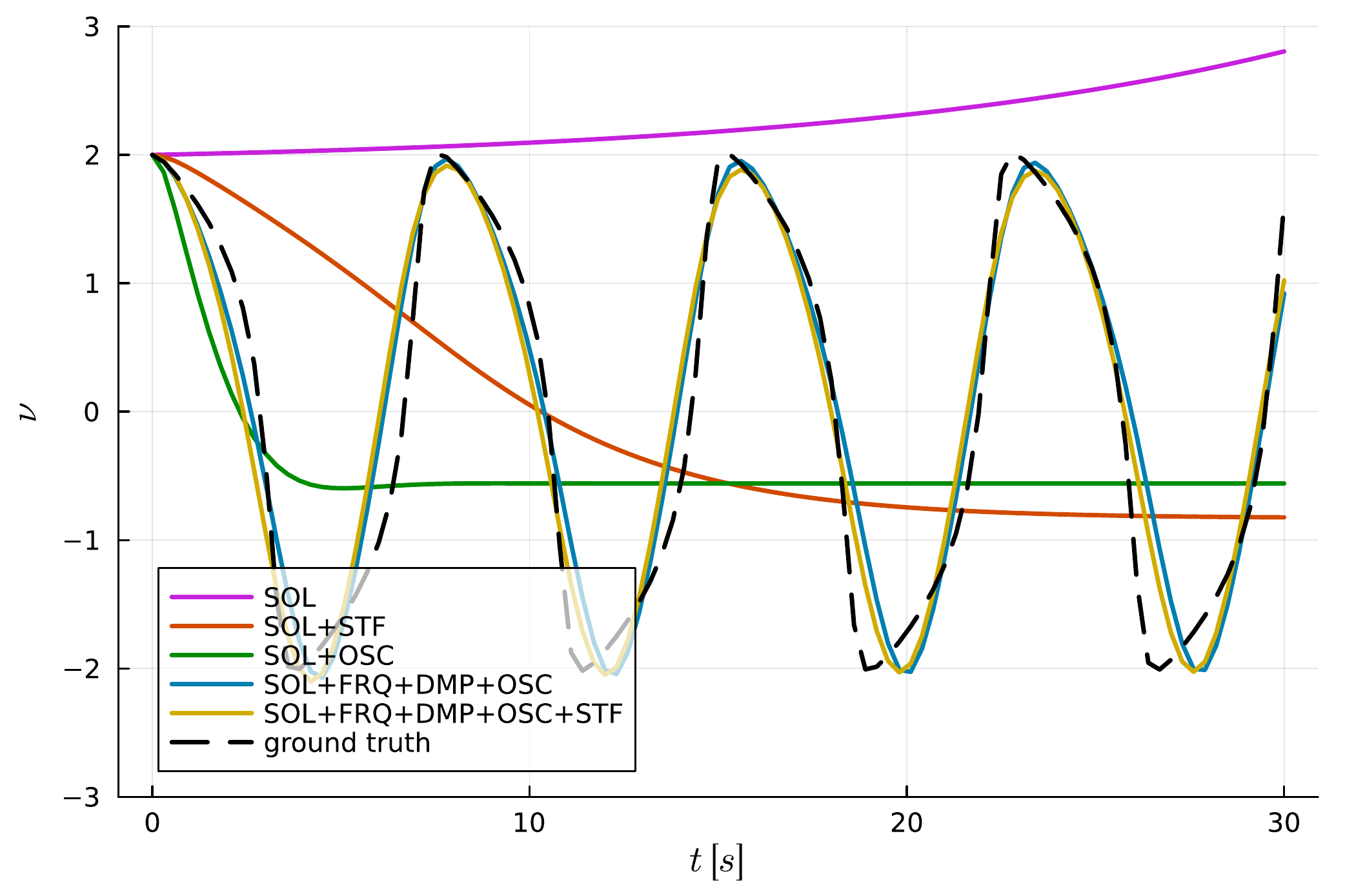}
	\caption{Comparison of the NeuralODE solutions ($\nu$), trained with different gradients. The gradient configuration SOL+FRQ+DMP+OSC (blue) and +STF (yellow) lay close to the ground truth (black-dashed). Further training will improve the fit.}
	\label{fig:transsol3_1}
\end{figure}

\begin{figure}[h!]
	\centering
	\includegraphics[width=\plotwidth]{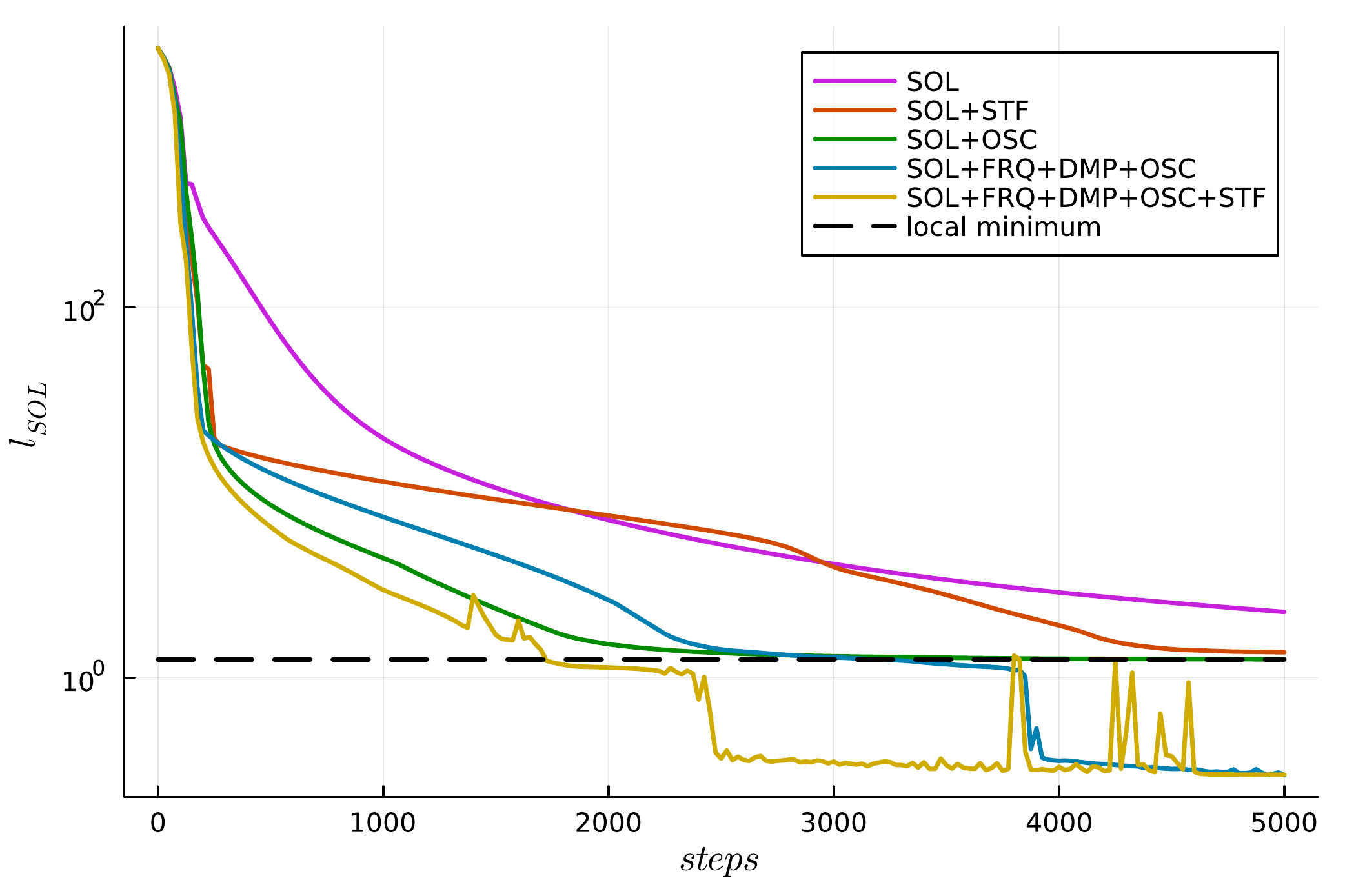}
	\caption{Comparison of the NeuralODE convergence behavior, trained with different gradients. Figure shows the loss defined on the ODE solution $l_{SOL}$. A strong local minimum is plotted black-dashed.}
	\label{fig:transconv3}
\end{figure}

As for the linear system examples, training on the solution gradient alone does not converge to the target solution (s. Fig. \ref{fig:transsol3_1}). Adding the STF or OSC gradient slightly improves the fit, but leads to convergence in a local minimum. The gradient configuration SOL+FRQ+DMP+OSC with and without STF is able to predict the behavior of the \ac{VPO}, further training and larger \ac{ANN} topologies will improve the fit. In analogy, convergence on basis of the loss function $l_{POS}$ can be observed in \reffig{transconv3}. In this example, adding the STF gradient (yellow) leads to faster convergence compared to the configuration without STF gradient (blue).

Concerning stability for this example is not purposeful, because the \ac{VPO} is not stable in its entire operating state space.
\FloatBarrier

\section{Conclusion}
We highlighted a suitable strategy to induce additional system knowledge in form of \ac{ODE} properties into a NeuralODE training process and obtained an \emph{eigen-informed NeuralODE}. From a technical view, \ac{ODE} properties are translated to eigenvalue positions. These positions of eigenvalues can be easily considered in the loss function design. To maintain a gradient based training for eigen-informed NeuralODEs, also the sensitivities over the eigenvalue operations are needed. We deploy a suitable implementation for this in the Julia programming language, a ready to use implementation of a differentiable eigenvalue and -vector function is available open source as \emph{DifferentiableEigen.jl} (\url{https://github.com/ThummeTo/DifferentiableEigen.jl}). 

Common NeuralODEs tend to converge to local minima and are not protected from becoming (unintentionally) instable. We showed that eigen-informed NeuralODEs are capable of avoiding these problems and outperform vanilla NeuralODEs in the presented disciplines. We further exemplified, that eigen-informed NeuralODEs are capable of handling modeling applications, that are unsolvable for common NeuralODEs: Eigen-informed NeuralODEs allow for training on basis of insufficient data, meaning data that does not fulfill the Nyquist-Shannon sampling theorem, if the system eigenmodes are known. 

Last but not least, knowledge of ODE properties like stability, frequencies, damping and/or stiffness can improve training convergence in many different use cases, regardless of issues from local minima or instability. The presented technique, the eigen-informed NeuralODE, can often be implemented for a moderate additional computational cost compared to the deployment of vanilla NeuralODEs, especially if implicit solvers are used. Whether the trade-off (additional computational cost, but faster convergence) is economical must be examined on a case-by-case basis. For problems that are not solvable at all by pure NeuralODEs, deploying the presented method will often be beneficial.

Basically, the presented method is not limited to the use in NeuralODEs, but opens up to the entire family of Neural Differential Equations. This for example includes Neural Partial Differential Equations that are often used for fluid dynamic simulations.

Future work covers a more detailed study of the different gradients and the extension of this concept to the use in physics-enhanced NeuralODEs \cite{Thummerer:2022}, the combination of a conventional ODE, \ac{ANN} and an \ac{ODE} solver. Besides dealing with instability and local minima, this will also open up to nonlinear control design. We will show, that eigen-informed physics-enhanced NeuralODEs can be easily used to train powerful nonlinear controllers, simply by defining a loss function on basis of a target trajectory, together with an additional gradient for pushing the instable eigenvalues to the left half of the complex plane - satisfying the universal controller design objective. Further, for training on large applications with many states and eigenvalues, a reliable eigenvalue tracking method is needed and an active topic of research.

In general, other eigenvalue methods, like eigenvalue computation via \ac{ANN} as in \cite{Yi:2004} might be interesting, because of the seamless compatibility with \ac{AD} and (possibly) improved computational performance. 

\printbibliography

\end{document}